\documentclass{article}

\UseRawInputEncoding 
\usepackage{PRIMEarxiv}

\usepackage[utf8]{inputenc} 
\usepackage[T1]{fontenc}    
\usepackage{hyperref}       
\usepackage{url}            
\usepackage{booktabs}       
\usepackage{amsfonts}       
\usepackage{nicefrac}       
\usepackage{microtype}      
\usepackage{marvosym}
\usepackage{lipsum}
\usepackage{fancyhdr}       
\usepackage{graphicx}       
\graphicspath{{media/}}     
\usepackage{algorithm}
\usepackage{algorithmic}
\usepackage{graphicx}     
\graphicspath{ {./images/} } 
\title{CADG: A Model Based on Cross Attention for Domain Generalization }
\author{
  Cheng Dai\textsuperscript{1}, Yingqiao Lin\textsuperscript{2}, Fan Li \textsuperscript{1*}, Xiyao Li\textsuperscript{3}\\
  Kuaishou Technology \\
  \texttt{\{daicheng, linyingqiao, lifan07, lixiyao\}@kuaishou.com} \\
   \And
  Donglin Xie \textsuperscript{4} \\
  Sichuan University \\
  \texttt{2020223045208@stu.scu.edu.cn} \\
}

\begin{document}
\maketitle

\begin{abstract}
In Domain Generalization (DG) tasks, models are trained by using only training data from the source domains to achieve generalization on an unseen target domain, this will suffer from the distribution shift problem. So it's important to learn a classifier to focus on the common representation which can be used to classify on multi-domains, so that this classifier can achieve a high performance on an unseen target domain as well. With the success of cross attention in various cross-modal tasks, we find that cross attention is a powerful mechanism to align the features come from different distributions. So we design a model named CADG (cross attention for domain generalization), wherein cross attention plays a important role, to address distribution shift problem. Such design makes the classifier can be adopted on multi-domains, so the classifier will generalize well on an unseen domain. Experiments show that our proposed method achieves state-of-the-art performance on a variety of domain generalization benchmarks compared with other single model and can even achieve a better performance than some ensemble-based methods. 
\end{abstract}


\section{Introduction}
Currently, deep learning models have achieved remarkable performance in a wide range of aspects in the independent and identically distributed (i.i.d) setting. But in a real-word application, distribution shift caused by physical or psychological factors can always be found, which breaks the i.i.d assumption and make the models suffer from poor generalization. So more and more studies has focus to address this problem and domain generalization\cite{blanchard2011generalizing}(DG) is one of the most popular and promising studies.

A data distribution can be seen as a domain and domain generalization (DG) aims to address distribution shift by learning a classifier that can generalize well on the test data sampled from an unseen distribution,  after training on more than one training domains. In order to improve the generalization on the test domain in DG setting, various methods have been proposed. Some recent works demonstrate that using empirical risk minimization\cite{vladimir1998statistical} (ERM) along with proper model selection (i.e. early stopping using validation set) under a fair evaluation protocol, called DomainBed \cite{gulrajani2020search}, can already make the classifier to achieve a competitive performance against most previous domain generalization methods which focus on learning domain-invariant feature representations. 

With the rising of DomainBed, more and more domain generalization methods focus on using some learning algorithms and model selection to achieve a better performance and one of the most popular and useful methods is ensemble learning\cite{hansen1990neural} \cite{zhou2018diverse}. This make the less seeking on the model design for addressing the distribution shift problem and will not be helpful to touch the boundary of the generalization with a single model. Besides, a ensemble method will cause more resource cost in application phrase and more work in update phrase, so training a end-to-end model to address the distribution shift problem will be more applicable. In this case, we come back to research a better way to align the distributions of source and target domains. 

Inspired by some other works involve the feature alignment such as cross-modal retrieval\cite{geigle2021retrieve} and multi-modal representation\cite{hu2021unit}, and with the success of the transformer used in CV aspect, we find that cross-attention is a powerful mechanism to align the feature come from different distributions. So we design a model named CADG (cross attention for domain generalization), wherein cross-attention play a important role, to address distribution shift problem. in our method, data come from different training domains with the same label will be organized as a data pair and we will use cross attention to extract the stable representation which can be used to classify from this pair. 

Our contributions can be summarized as follow:
1. We introduce a new feature alignment way with cross attention to extract the stable representation which can be used to classify for domain generalization. As the model is based on transformer, so various performance improvement methods can be introduced in our method in future work.
2. Our proposed method achieves state-of-the-art performance on a variety of domain generalization benchmarks compared with other single models and can even outperform some ensemble-based methods.

\section{Related work}
\label{sec:headings}

\subsection{Domain generalization}
As \cite{wang2021generalizing} summarized, most existing domain generalization methods can be divided into following categories.
(1) Data manipulation. This idea is first proposed in \cite{tobin2017domain}, which aims to create diverse training data to assist learning general representations on unseen target domain.  Along this line, \cite{peng2018sim} and \cite{tremblay2018training} improve the generalization of the models through domain randomization and some other works such as \cite{kim2021selfreg} use self-supervised learning to  strengthen the generalization capability of the models.
(2) Representation learning. This category of methods aims to find the invariant representation across various domains and one of the most popular techniques is domain alignment, which usually use some  regularizations to align feature across domains explicitly such as maximum mean discrepancy \cite{pan2010domain} (MMD), second order correlation \cite{sun2016return} and Wasserstein distance \cite{zhou2020domain}. Our method belongs to representation learning but is a implicit way via cross-attention.
(3) Learning strategy: This category of methods focuses on exploiting the general learning strategy to promote the generalization capability. Ensemble learning is one of the most available methods such as \cite{cha2021swad}\cite{arpit2021ensemble}, but as we discussed before, these methods are not very friendly in a real-word application.

\subsection{Transformer}
Transformer\cite{vaswani2017attention} is first proposed to model sequential data in NLP tasks and has been proved excellent performance. In recent years, with the effectiveness of transformer has been showed in  computer-vision tasks\cite{han2020survey}. Through feeding transformer with sequences of image patches, such a VIT\cite{dosovitskiy2020image} and many other ViT variants\cite{yuan2021tokens}\cite{liu2021swin}, more and more studies have focused on using transformer based models to address CV tasks, cross-modal retrieval tasks and multi-modal representation tasks. Attention mechanism is the core of transformer and a extension of it named cross-attention has been proved very powerful at distilling noise and feature alignment in several works\cite{hu2021unit}\cite{tsai2019multimodal} for multi-modal based networks. This inspires us than if we regard the data from different modals as a special distribution, the situation that we should align features from different distributions in multi-modal tasks will be very similar with domain generalization tasks. So in this paper, we apply cross-attention to align feature across multi-domains to address the distribution shift problem.

\section{The Proposed Method}
We first introduce our proposed co-attention module and analyze its role in eliminating domain differences. Then the two-way center-aware training method is presented in Section 3.2 . With the constructed pair as sample input, our CDAG  is proposed in Section 3.3,  which consists of four weight-sharing transformers.
\subsection{The Transformer Based Vision Cross Attention}
In many computer vision tasks, vision transformer (ViT \cite{dosovitskiy2020image} )  has achieved comparable or even superior performance. The self-attention(\cite{vaswani2017attention}) module of ViT is one of the most important structure. ViT reshapes an image $I \in  I^{H \times W \times C}$ into a sequence of flattened 2D patches $x \in R^{N \times (P^{2}\cdot C)}$, where (H, W) is the height and width of the origin image, C is the channel numbers, P is the resolution of each image patch. Therefore, we can get a total of $N = H \cdot W/P^2$  patches. In self-attention, the image patches are first mapped into three vectors by projected, which are queries $Q \in R^{N \times  d_q}$, keys $ K \in R^{N \times d_k}$, and value $ V \in R^{N\times d_v}$. The output of self-attention is obtained by weighted sum of the value, where the weight for each value is computed by the query and key, through a compatibility function. The self-attention module takes N patches as input and aims to mine the relationship between the patches of the input image. Its  calculation process is as follows:
$$SelfAttn(Q,K,V) = softmax(\frac {QK^T}{\sqrt{d_k}})V$$

The difference between self-attention and cross-attention if that the input of cross-attention is an images pair, such as $I_1, I_2$. The query  of cross-attention is from $I_1$, while the key/value is from $I2$. Its  calculation process is:
$$CrossAttn(Q_1, K_2,V_2) = softmax(\frac {Q_1K_2^T}{\sqrt{d_k}})V_2$$

The M patches from $I_1$ consist of the queries $Q_1 \in R^{M \times d_k}$ ,while the $K_2 \in R^{N \times d_k}$ and $V_2 \in R^{N \times d_v}$ are keys and values from N patches of image $I_2$. Cross-attention has M outputs, each output is a weighted sum of $V_t$, and the weights are calculated by a compatibility function from the queries of $I_1$ and the keys in $I_2$. The result of cross-attention is that the path of $I_2$ will have a greater weight if it's more similar to query of $I_1$.

The input of cross-attention is an image pair from two different domains. By means of cross-attention, the alignment of different domains is realized in the latent space, and the difference between domains is eliminated. Specifically, for two images from different domains, split them into patches and perform cross-attention. Similar patches from different sources have greater attention weights than dissimilar patches. In this way, similar attributes between different domain data are preserved, and dissimilar ones are eliminated, so as to eliminate the differences between domains and extract domain invariant features.

\begin{figure}
\centering   
\includegraphics [scale=0.2]{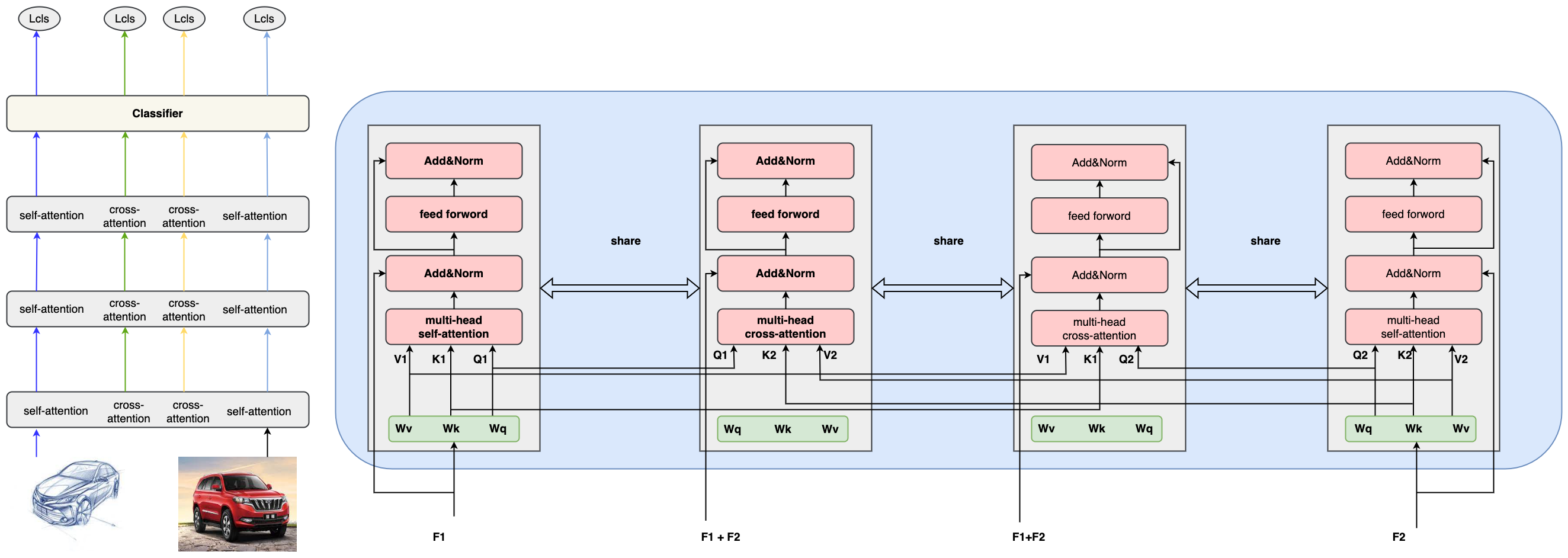}  
\caption{domain-generation transformer layer} 
 \label{fig:1} 
\end{figure}

\subsection{Pair-wised Training Process}
In general domain generalization research, data from multiple domains are mixed together, the model inputs a picture, and the model is monocentric. In this paper, we propose a pair-wised  training method. First, we sample a category from the set of categories in the sample, then sample two domains from multiple training domains, and sample images from the corresponding categories of these two domains to construct a picture pair. Each image is fed into the model for self-attention, and then the image pair is fed into the model for cross-attention. The detailed algorithm flow is described in algorithm \ref{algorithm1}.
\begin{algorithm} 
	\caption{Two-Way Center-Aware Training} 
	\label{alg3} 
	\begin{algorithmic}
		\STATE Initialization:$ D = {D_1, D_2, D_3,...D_m} $  are  m  domains
		\STATE Initialization:$ (x_i^j, y_i^j) $ is the j-th sample of domain $D_i$
		\STATE Initialization:$C = {C_1, C_2, ..., C_k}$ are k classes, $epoch$ is the number of epochs, $S$ is the steps for an epoch.
		\STATE Initialization:$model$, $e \gets 0$, $s \gets 0$
		\WHILE{$e <  epoch$} 
		    \WHILE {$s < S$} 
		        \STATE sample one class $C_r$ from C 
		        \STATE sample two domains $D_p, D_q$ from D
		        \STATE sample  a data $(x_p^i, y_p^i)$ from $D_p$ and  another data $(x_q^j, y_q^j)$ from $D_q$, where $y_p^i = C_r$ and $y_q^j = C_r$
		        \STATE $ loss = model(x_p^i, x_q^j)$
		        \STATE $ backward(loss)$
		        \STATE $ s \gets s + 1 $
		    \ENDWHILE
		     \STATE $ e \gets e + 1 $
		\ENDWHILE 
	\end{algorithmic} 
	\label{algorithm1}
\end{algorithm}

\subsection{CADG: Cross Attention for Domain Generalization }
The framework of the proposed method is shown in Figure \ref {fig:1}. Our model consists of four
weight-sharing transformer blocks. There are four data flows and constraints for the weight-sharing branches. It consists of 4-way weight-sharing transformer modules, two of which are self-attention modules named as self-branch1 and self-branch2. The others are cross-attention modules named as cross-branch1 and cross-branch2. The input to the framework is two images with the same label from two different domains. As shown in Fig\ref{fig:1}, the two images are input to two self-attention branches for learning domain-specific features. The outputs of the two branches are respectively connected to the softmax classifier for supervised classification training.

The inputs of the cross-attention come from the outputs of the two self-attention modules. For the N-th layer, the queries of the cross-branch1 module come from the queries of the N-th layer's self-branch1, and the keys and values come from the corresponding self-branch2. The inputs of cross-branch2 is just the opposite of cross-branch1, which its queries come from the queries of the N-th layer's self-branch2, while the  keys and values come from the corresponding self-branch1. The outputs of the cross-attention modules are added to the outputs of the (N-1)-th layer.

Each of the four branches will access a softmax classifier with shared parameters and calculate the cross entropy loss. The four losses are weighted to obtain the final classification loss.
$$ L_{cls} = \lambda_{1} loss_1 + \lambda_{2} loss_2 + \lambda_{3} loss_{cross1} + \lambda_{4} loss_{cross2} $$
where $loss_1$ and $loss_2$ are the loss  of the two self-attention branches, $loss_{cross1}$ and $loss_{cross2}$ are the cross-attention branch's losses.

\section{Experiment}
\subsection{Datasets and Implementation}
The proposed method is verified  on 5 real-world benchmark datasets including PACS (4 domains, 9,991 samples, 7 classes) \cite{li2017deeper}, VLCS (4 domains, 10,729 samples, 5 classes) \cite{torralba2011unbiased}, OfficeHome (4 domains, 15,588 samples, 65 classes) \cite{venkateswara2017deep}, TerraIncognita (4 domains, 24,778 samples, 10 classes) \cite{beery2018recognition}, and DomainNet (6 domains, 586,575 samples, 345 classes) \cite{peng2019moment}.

We follow DomainBed's \cite{gulrajani2020search} training and validation protocols for fair comparisons. For model selection, we use the training domain validation set protocol. Specifically, the data of a certain domain is selected as the target domain, the rest are the source domain, and 20\% of the source domain is divided as the validation set. We run training and validation 3 times, using different random seeds, with different splits of the training-validation set. The out-of-domain test performance averaged over all domains will be reported for each dataset. Except for DomainNet\cite{peng2019moment} which is 15000 steps, we use the standard 5000 iterations for other datasets, and reduce unnecessary computation by early-stop based on the validation set accuracy.

This section presents experimental results on the DomainBed suite. The input image size in our experiments is 224×224. We user the DeiT-base \cite{touvron2021training} which is pretrained on  ImageNet1K as our backbone. We use the Stochastic Gradient Descent algorithm with the momentum of 0.9 and weight decay ratio 1e-4 to optimize the training process. The learning rate is set to 3e-3 for all datasets. The batch size is set to 64. 

Firstly, our algorithm is compared with a non-ensemble algorithm, and the results are shown in Table \ref{Table1}. It can be found that our proposed method, without any tricks, outperforms all non-ensemble algorithms and reaches the sota level. In all experiments, our proposed method has significant improvement over non-ensemble methods: +3.1pp in PACS, +0.9pp in VLCS, +11.3pp in OfficeHome, +4.9pp in TerraIncognita and +7.1pp in DomainNet. On all datasets, our method has an overall average improvement of 6.6 pp compared to the non-ensemble algorithm.

We also compare our algorithm with the ensemble algorithm which are pretrained on ImageNet, and the results are shown in Table \ref{Table2}. It can be found that without any ensemble, our algorithm can surpass all ensemble algorithms on the list and become state-of-the-art with 1.1pp improvement. Comparing the best results of the ensemble algorithm, our algorithm  outperforms on the two datasets: +0.5pp  on PACS,  +2.4pp on DomainNet.

\begin{table}
    \centering
    \begin{tabular}{|l|l|l|l|l|l|l|}
    \hline
        Algorithm & PACS & VLCS & OfficeHome & TerraIncognita & DomainNet & avg \\ \hline
        DANN (JMLR’16) \cite{ganin2016domain}  & 84.6\textpm 1.1 & 78.7\textpm 0.3 & 65.4\textpm 0.6 & 48.4\textpm 0.5 & 38.4\textpm 0.0 & 63.1 \\ \hline
        CORAL (ECCV’16)\cite{peng2018synthetic} & 86.0\textpm 0.2 & 77.7\textpm 0.5 & 68.6\textpm 0.4 & 46.4\textpm 0.8 & 41.8\textpm 0.2 & 64.1\\ \hline
        MMD (CVPR’18) \cite{li2018domain} & 85.0\textpm 0.2 & 76.7\textpm 0.9 & 67.7\textpm 0.1 & 49.3\textpm 1.4 & 39.4\textpm 0.8 & 63.6 \\ \hline
        C-DANN (ECCV’18) \cite{li2018deep} & 82.8 \textpm 0.5 & 78.2\textpm0.4 & 65.6\textpm 0.5 &47.6\textpm0.8 & 38.9\textpm 0.1 & 62.6\\ \hline
        ERM (ICLR’21) \cite{gulrajani2020search} & 85.7\textpm 0.5 & 77.4\textpm 0.3 & 67.5\textpm 0.5 & 47.2\textpm 0.4 & 41.2\textpm 0.2 & 63.8 \\ \hline
        Fishr (ICML’22)\cite{rame2021fishr}  & 85.5\textpm 0.4 & 77.8\textpm 0.1 & 67.8\textpm 0.1 & 47.4\textpm 1.6 & 41.7\textpm 0.0 & 67.1 \\ \hline
        CADG (ours)& \textbf{94.6 \textpm 0.42} & \textbf{82.2 \textpm 1.1} & \textbf{79.9 \textpm 0.5} & \textbf{55.7 \textpm 0.2} & \textbf{51.6 \textpm 0.0} & \textbf{72.7} \\ \hline
    \end{tabular}
    \caption{Non-ensemble algorithms}
    \label{Table1}
\end{table}

\begin{table}
    \centering
    \begin{tabular}{|l|l|l|l|l|l|l|}
    \hline
        Algorithm & PACS & VLCS & OfficeHome & TerraIncognita & DomainNet & avg \\ \hline
        SWAD (NIPS’21) \cite{cha2021swad} & 88.1\textpm 0.1 & 79.1\textpm 0.1 & 70.6\textpm 0.2 & 50.0\textpm 0.3 & 46.5\textpm 0.1 & 66.9 \\ \hline
        EoA (arxiv) \cite{arpit2021ensemble} & 88.6 & 79.1 & 72.5 & 52.3 & 47.4 & 68.0\\ \hline
        random ensemble  & 58.1\textpm 0.13 & 58.5\textpm 1.26 & 59.6\textpm 0.38 & 31.5\textpm 0.4 & 15.8\textpm 1.4 & 44.5 \\ \hline
        SEDGE\cite{li2022domain} & 84.1\textpm 0.45 & \textbf{79.8\textpm 0.12} & \textbf{79.9\textpm 0.12} & \textbf{56.8\textpm 0.21} & 46.3\textpm 0.39 & 69.4 \\ \hline
         CADG(ours) & \textbf{89.1 \textpm 0.3} & 79.6 \textpm 0.2 & 79.9 \textpm 0.0 & 54.2 \textpm 0.5 &\textbf{49.8 \textpm 0.1} & \textbf{70.5} \\ \hline
    \end{tabular}
    \caption{ensemble algorithms which pretrained on ImageNet}
    \label{Table2}
\end{table}

Although model pretrained on ImageNet is in common use, Kumar \cite{kumar2022fine} finds that model pretrained on
ImageNet may not be good for all datasets such as DomainNet. By using model that  pretrained on the CLIP dataset \cite{radford2021learning}, our model further improves the average performance by 2.2 pp .The details can be found in table \ref{Table3}.

\begin{table}[!ht]
    \centering
    \begin{tabular}{|l|l|l|l|l|l|l|}
    \hline
        Algorithm & PACS & VLCS & OfficeHome & TerraIncognita & DomainNet & avg \\ \hline
        EoA$^+$ (arxiv) \cite{arpit2021ensemble} & 93.2 & 80.4 & 80.2 & 55.2 & 54.6 & 72.7\\ \hline
        random ensemble  & 59.5\textpm 0.5 & 61.1\textpm 0.12 & 59.5\textpm 0.07 & 30.8\textpm 0.37 & 18.7\textpm 0.62 & 46.0 \\ \hline
        SEDGE$^+$ \cite{li2022domain}  & \textbf{96.1\textpm 0.04} & 82.2\textpm 0.03 & \textbf{80.7\textpm 0.21} & \textbf{56.8\textpm 0.29} & \textbf{54.7\textpm 0.1} & \textbf{74.1} \\ \hline
        CADG$^+$ (ours)& 94.6 \textpm 0.42 & \textbf{82.2 \textpm 1.1} & 79.5 \textpm 0.5 & 55.7 \textpm 0.2 & 51.6 \textpm 0.0 & 72.7 \\ \hline
    \end{tabular}
    \caption{ensemble algorithms which pretrained on Clip}
    \label{Table3}
\end{table}

\section{Conclusions}
 In this paper, we propose a single-model domain generalization algorithm based on cross-attention transformer, which can achieve feature alignment and extract invariant features. It does not require the use of ensemble methods, that greatly facilitates the application of such algorithms in practical scenarios. Based on this paper, subsequent work related to transformers can also be introduced into the task of domain generalization. Experiments on five benchmark datasets show that our proposed method achieves state-of-the-art performance compared with other single model and can even achieve a better performance than some ensemble-based methods. 

\bibliographystyle{IEEEtran}
\bibliography{references}

\end{document}